\documentclass[sigconf]{acmart} 

\AtBeginDocument{%
  }

\renewcommand\footnotetextcopyrightpermission[1]{} 
\copyrightyear{2024} 
\acmYear{2024} 
\setcopyright{rightsretained} 
\acmConference[DAC '24]{61st ACM/IEEE Design Automation Conference}{June
 23--27, 2024}{San Francisco, CA, USA}
\acmBooktitle{61st ACM/IEEE Design Automation Conference (DAC '24), June
 23--27, 2024, San Francisco, CA, USA}
\acmDOI{10.1145/3649329.3657350}
\acmISBN{979-8-4007-0601-1/24/06}

\acmSubmissionID{24}
\begin{document}

\title[Explainable FNN with MFRL for $\mu$-arch DSE]{Explainable Fuzzy Neural Network with \\Multi-Fidelity Reinforcement Learning for \\Micro-Architecture Design Space Exploration}

\author{Hanwei Fan$^\dag$,\; Ya Wang$^\dag$,\; Sicheng Li$^\mathsection$,\; Tingyuan Liang$^\dag$,\; Wei Zhang$^\ddag$}
\affiliation{
  \institution{$^\dag$ $^\ddag$ The Hong Kong University of Science and Technology, Hong Kong SAR, China}
  \institution{$^\mathsection$ Alibaba Group (United States), Sunnyvale, California, United States}
  \country{$^\dag$\{hfanah, ywangmu, tliang\}@connect.ust.hk,\quad $^\mathsection$ sicheng.li@alibaba-inc.com,\quad $^\ddag$ wei.zhang@ust.hk}}

\renewcommand{\shortauthors}{Hanwei FAN et al.}
\renewcommand{\thefootnote}{}

\begin{abstract}
With the continuous advancement of processors, modern micro-architecture designs have become increasingly complex. The vast design space presents significant challenges for human designers, making design space exploration (DSE) algorithms a significant tool for $\mu$-arch design. In recent years, efforts have been made in the development of DSE algorithms, and promising results have been achieved. However, the existing DSE algorithms, e.g., Bayesian Optimization and ensemble learning, suffer from poor interpretability, hindering designers' understanding of the decision-making process. 
To address this limitation, we propose utilizing Fuzzy Neural Networks to induce and summarize knowledge and insights from the DSE process, enhancing interpretability and controllability. 
Furthermore, to improve efficiency, we introduce a multi-fidelity reinforcement learning approach, which primarily conducts exploration using cheap but less precise data, thereby substantially diminishing the reliance on costly data.
Experimental results show that our method achieves excellent results with a very limited sample budget and successfully surpasses the current state-of-the-art. Our DSE framework is open-sourced and available at https://github.com/fanhanwei/FNN\_MFRL\_ArchDSE/\ .
\end{abstract}

\maketitle

\footnotetext{$^\ddag$ Corresponding author\\
* This work was partially supported by Hong Kong Research Grants Council General Research Fund (Grant No. 16213422).
}

\section{Introduction}
In the modern era, processors are indispensable, handling diverse workloads. To achieve optimal performance across varying application scenarios, processors require different micro-architecture ($\mu$-arch) configurations. However, the huge design space poses significant challenges for human designers to conduct design space exploration (DSE) manually.  
In recent years, researchers have tried various approaches to promote the use of automatic DSE algorithms to replace manual $\mu$-archs configuration tuning. Early work \cite{ipek2006efficiently, lee2007illustrative} proposed the classic $\mu$-archs DSE framework that combines statistical sampling and regression model. This kind of method randomly chooses a small number of representative samples to fit a regression model that can quickly predict the design metrics and then selects the most promising designs based on the results from the regression model, thereby reducing the number of samples that need to be examined and improving the efficiency of $\mu$-archs DSE. Subsequently, ActBoost\cite{li2016efficient} improves this framework by using Adaboost as the regression model to obtain more accurate predictions of the metrics while also using active learning to improve the sampling efficiency. More recently, Boom-Explorer\cite{bai2021boom} proposes to use Bayesian optimization (BO) with deep kernel-based Gaussian Process \cite{wilson2016deep} to solve $\mu$-archs DSE tasks, achieving state-of-the-art results with high sample efficiency. Also, \cite{wang2023high} proposes using bagging-based GBRT as the regression model and achieves excellent results. 

However, the existing $\mu$-arch DSE algorithms lack interpretability, making it difficult for designers to understand the rationale behind the algorithm's decisions, limiting their ability to derive insights or maintain control over these algorithms. On one hand, the algorithms' accumulated experience and knowledge during the DSE process are hard to visualize and interpret, making it difficult for designers to reference when optimizing designs further.
On the other hand, the inherent randomness and black-box nature of these algorithms make their behavior unpredictable, hindering designers from adjusting the algorithm to specific search requirements.
Therefore, there is a pressing need to develop DSE algorithms that are more interpretable and user-friendly for human designers.

\begin{figure}[t]
\centerline{\includegraphics[width=0.8\columnwidth]{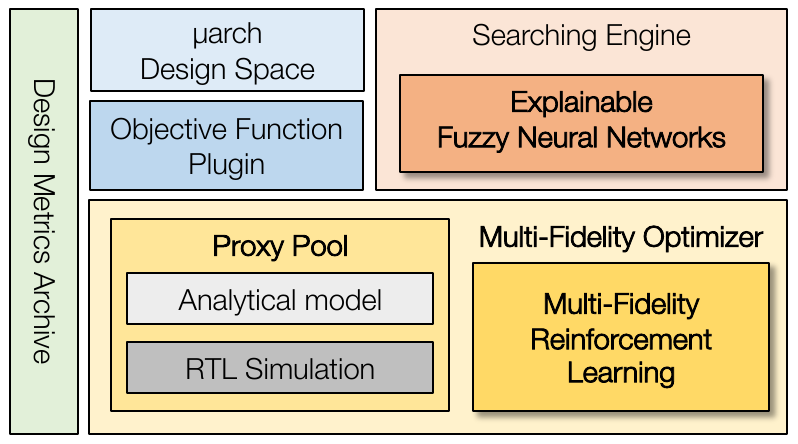}}
\caption{The framework of our proposed methods.}
\label{fig.frame}
\vspace{-0.6cm}
\end{figure}

 To address the interpretability issue, fuzzy rule-based DSE algorithms have been developed \cite{zeng2010framework, farhat2016fuzzy}. These algorithms, built on the observation that human designers use their knowledge and experience in computer architecture to optimize designs, embed such expertise into a set of fuzzy rules that guide the DSE process. This results in a rule-based system that achieves competitive DSE results while remaining a white-box model for designers. 
However, as the design space grows, the number of rules increases exponentially, leading to high costs in building the rule base. Additionally, designing effective rules requires expertise in fuzzy logic systems, which might be challenging for hardware designers. As a result, fuzzy rule-based DSE algorithms are rarely adopted.
To solve this problem, we propose the use of Fuzzy Neural Networks (FNN) \cite{jang1993anfis}, which is a class of artificial neural networks that incorporate fuzzy logic. The FNNs can be trained with Reinforcement Learning (RL) to obtain the design rules automatically \cite{lin1994reinforcement} so that the practicality of fuzzy rule-based DSE algorithms is enhanced.

Furthermore, existing DSE algorithms typically use a single proxy to evaluate design metrics of $\mu$-archs, which can either result in inaccurate outcomes or lead to a time-consuming DSE process. Commonly used evaluation proxies include analytical models \cite{jongerius2017analytic} and RTL simulators \cite{synopsys2004verilog}. Analytical models employ mathematical formulas to evaluate design metrics, offering high computational efficiency and rapid assessment speed. However, their inherent simplifying assumptions and high level of abstraction from the actual architecture often compromise accuracy. On the other hand, RTL simulators, software tools that simulate the behavior of a processor cycle-by-cycle, provide highly accurate estimations but at the expense of large time overhead. In practice, designers usually use fast analytical models to locate the regions of interest in the large design space to save time, then conduct fine-grained tuning using RTL simulators to ensure accurate results. This common practice inspires us to develop a multi-fidelity RL algorithm that incorporates insights from both the analytical model and RTL simulation, with the aim of achieving accurate results while significantly reducing time consumption. As an imitation of the $\mu$-archs tuning process of human designers, the combined FNN and multi-fidelity RL DSE framework maintains excellent interpretability.

The framework we propose is illustrated in Fig.\ref{fig.frame}. We would like to highlight the following contributions:
\begin{itemize}
\item We propose to adopt FNN as the search engine for decision-making in the DSE process. This approach can autonomously formulate design rules encapsulating the insights and experience acquired during exploration. To the best of our knowledge, this is the first attempt to utilize FNN for explainable $\mu$-archs DSE.
\item We develop a multi-fidelity RL algorithm to train the FNN, which uses both the analytical model and RTL simulation to improve the efficiency of the DSE process significantly while guaranteeing the accuracy of the DSE results.
\item We conduct comprehensive experiments, showing that our DSE framework significantly outperforms the state-of-the-art DSE algorithms and enjoys good interpretability.
\end{itemize}

\section{Fuzzy Neural Networks for Micro-Architecture DSE}\label{sec.FNN}
The FNN is a hybrid model that combines the principles of fuzzy logic and the structure of neural networks. It is a powerful tool that takes advantage of both numerical and linguistic information to solve complex problems. In this section, we introduce the basics of the FNN and how we apply it to $\mu$-arch DSE.

\subsection{Fuzzy Logic}
Fuzzy logic \cite{zadeh1988fuzzy} employs fuzzy rules to describe the relationships between variables. These fuzzy rules are structured as if-then statements. For example, an instance of such a rule could be "if cycle per instruction (CPI) is `high` and cache size is `small`, then the cache set number should be `increased`". In this context, the 'if' part (e.g., CPI is high and cache size is small) is known as the antecedent, and the 'then' part (e.g., cache set number should be increased) is referred to as the consequent. The adjectives used (high, small, increase) are known as fuzzy variables.

Formally, a fuzzy rule \(R_i\) can be written as:
\[ R_i: \text{{IF }} x_1 \text{{ IS }} A_{i1} \text{{ AND }} \ldots \text{{ AND }} x_n \text{{ IS }} A_{in} \text{{ THEN }} y \text{{ IS }} B_i \]
where \(x_1, \ldots, x_n\) are the antecedents, \(y\) is the consequent, and \(A_{i1}, \ldots, A_{in}, B_i\) are fuzzy sets.

Fuzzy variables abstract the numerical values into more understandable terms, offering a user-friendly interface between the rules and the users. This makes the rules particularly suitable for encapsulating the knowledge and experience of human designers.
The transformation between crisp (numerical) values and fuzzy values is performed by membership functions (MFs). These MFs are mathematical functions that can take various forms, such as Sigmoid, Gaussian, and Bell functions. The transformation process, known as fuzzification, calculates the degree of membership (\(\mu\)) of each crisp value to the fuzzy sets. The degree of membership, ranging from 0 to 1, represents the extent to which a crisp value belongs to a fuzzy set. Notably, a crisp value can belong to multiple fuzzy sets simultaneously but with different \(\mu\). Formally, \(\mu\) of a crisp value \(x\) to a fuzzy set \(A\) can be calculated as $\mu_A(x)$.

\begin{figure}[tbp]
\centerline{\includegraphics[width=\columnwidth]{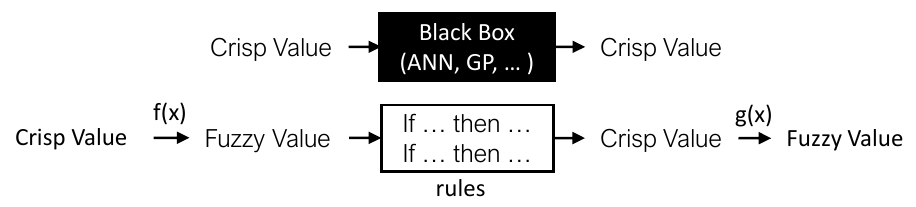}}
\caption{Comparison between black-box methods and Fuzzy rule-based system.}
\label{fig.fuzzycmp}
\vspace{-0.5cm}
\end{figure}

The ruling process activates the rules that contain the used fuzzy sets and typically uses a t-norm operator (such as the min or product operator) to calculate the firing strength of a rule \(R_i\), given by:
\begin{equation}
 \mu_{R_i}(x_1, \ldots, x_n) = T(\mu_{A_{i1}}(x_1), \ldots, \mu_{A_{in}}(x_n))
\end{equation}
where \(T\) is a t-norm operator.

Finally, the defuzzification process converts the fuzzy results back into crisp values, which are the consequences. Then, the output is the weighted average of results from all activated rules, which is represented as:
\begin{equation}
    y = \frac{\sum_{i=1}^n \mu_{R_i}(x_1, \ldots, x_n) \cdot y_i}{\sum_{i=1}^n \mu_{R_i}(x_1, \ldots, x_n)}
\end{equation}
where \(y_i\) are the crisp values of the rules.

As illustrated in Fig. 2, the bidirectional transformation process of fuzzy logic enables decision-making in the interpretable natural language form, while black-box methods operate exclusively with crisp values and lack transparency.

\subsection{Fuzzy Neural Networks}
\begin{figure}[tbp]
\centerline{\includegraphics[width=\columnwidth]{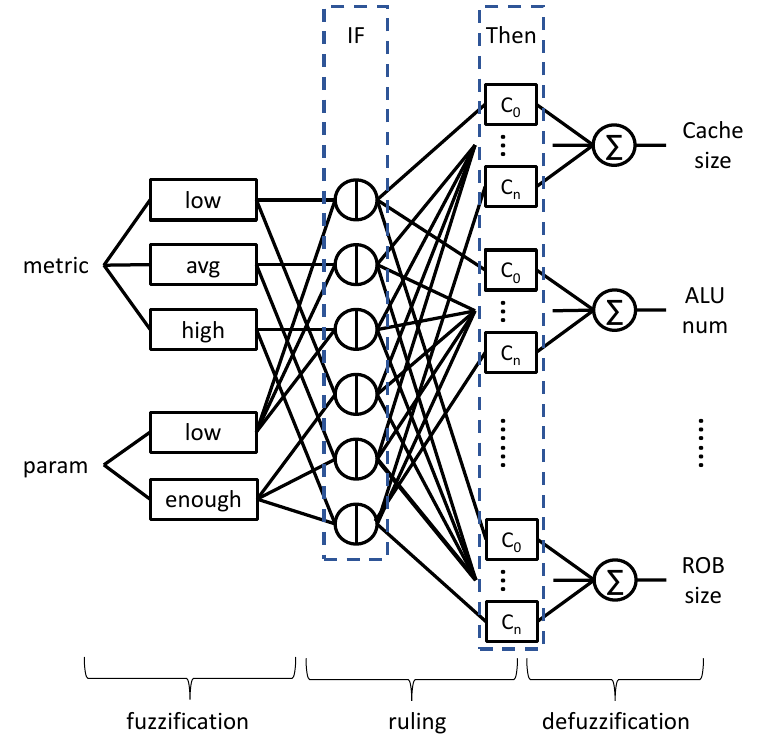}}
\caption{Structure of Fuzzy Neural Networks}
\label{fig.FNN}
\end{figure}

Despite the benefits brought by fuzzy logic, this method is not widely adopted for DSE due to the difficulty in building the fuzzy rule base. Therefore, it is important to automate the rule-making process. Fortunately, fuzzy logic shares very similar computation patterns with neural networks and thus can be formulated into FNNs. Fig. \ref{fig.FNN} shows the structure of FNNs, which implement the fuzzy logic process through five distinct layers:

\begin{itemize}
\item \textbf{Fuzzification Layer} takes design metrics and the current parameters as inputs and calculates the MFs.
\item \textbf{Ruling Layer} calculates the product of all the  \(\mu\) of the fuzzy values contained by each antecedent and outputs the firing strength of the rules.
\item \textbf{Normalization Layer} normalizes the rule's firing strength to ensure they are of a reasonable scale.
\item \textbf{Defuzzification Layer} defuzzifies the fuzzy values into crisp values. To simplify the computation, we adopt the Takagi-Sugeno (TS) \cite{takagi1983derivation} type fuzzy rules, where the consequent fuzzy value is directly represented by a crisp value. For instance, 'increase' can be represented by $C > 0$.
\item \textbf{Output Layer} returns the sum of the consequences weighted by firing strengths of the rules.
\end{itemize}

The weights of the FNNs have two parts, one of them being the consequent crisp values. The other part of them is the hyperparameters of the MFs, e.g., the center of Sigmoid and Bell. These hyperparameters represent the range of each fuzzy value.
For instance, if 'CPI high' uses the Sigmoid function as its MF and the center of the Sigmoid function is 5, this implies that a CPI value above 5 is considered 'high'. On the other hand, if the 'CPI avg' uses the Bell function as its MF and the center of the Bell function is 3, this suggests that a CPI value around 3 is considered 'average'.

As the entire FNN is differentiable, its weights can be updated by gradient descent, leading to the desired automatic rule-making.

\subsection{Adaptation for Micro-Architecture DSE}
In order to make FNN applicable to DSE, we made a series of adjustments. The input design metrics are categorized as 'low', 'avg', and 'high' with corresponding MFs: Inverse Sigmoid, Bell, and Sigmoid. 
The input design parameters are only categorized as 'low' and 'enough', with Inverse Sigmoid and Sigmoid MFs, respectively. 
The centers of these MFs can be defined by equally dividing the metric scale or using custom settings for faster convergence. 
However, drastic changes in the centers can activate different rules, rendering previous training ineffective. To avoid this, we disallow backpropagation for the centers of design metrics,  which are prone to substantial changes during gradient descent. 
However, the centers of the input design parameters are automatically updated to encourage better coverage as the mathematical properties of FNNs moderate their variations.

The antecedent of each rule contains all the inputs of the FNN, and all combinations of antecedents will have a corresponding rule. Therefore, the number of rules is as large as $3^{\# metrics} *\ 2^{\# params}$. To enhance efficiency and facilitate inspection, we can merge related design parameters, e.g., merge cache set and way as cache size.

The outputs of the FNN are the scores for all design parameters. In our DSE setting, the initial design is the smallest $\mu$-arch in the design space, and at each step the parameter with the highest score from the FNN is increased by 1. Thus, the 'THEN' part of the rule is translated as 'The parameter with the highest score should increase'.

Based on the interpretability of our proposed FNN, the designers can easily inspect the training results and take control of the training.
Firstly, when the training doesn't converge well, users can check on the rules and find the abnormal patterns, based on which the training setting can be easily adjusted. 
For example, if the centers of the MFs are updated beyond the limits of the design space, we can infer that the learning rate needs to be reduced. Furthermore, if a rule indicates that a design parameter should increase even when it's already at a high value, we can adjust the design space to concentrate on the higher range of this parameter. 

Secondly, to accelerate the training, the centers of design parameters can be wisely initialized based on the obvious features of target applications. For example, if the application has a large data size, the center of the cache size can be given a higher value.

In addition, the FNN allows us to incorporate our preferences directly into the rule base. For example, if we wish to favor designs with a decode width of 4, we can define 3 as 'low' and 4 as 'enough' in the antecedent part of the rule. We then adjust the corresponding consequence to increase the decode width when it falls short. These features enhance the flexibility and usability of the FNN, setting it apart from black-box methodologies. Empirical evidence supporting these benefits will be presented in Section \ref{sec.exp}.

\section{Multi-Fidelity Reinforcement Learning}
Considering designers often need to optimize processor performance within limited chip areas in real-life chip design scenarios, the goal of our proposed algorithm is to minimize the cycle per instruction (CPI) metric with a given constraint on the area. In each episode, we enlarge the processor step by step until the area limit is reached so that all the sampled designs are valid. For each step, we randomly choose one design parameter to increase and evaluate the area with a rapid area model. The CPI of the final design of an episode is the reward of all actions in this episode, and we update the FNN using policy gradient \cite{sutton1999policy}.

Based on this RL setting, training an FNN will consume a large number of samples whose design metrics need to be evaluated. 
To enable agile development, designers usually first conduct DSE on computationally efficient analytical models to find the promising area of the design space. Then, the designers can further use the HF simulations to perform local search in the narrower space. Such a design process is desired to be automated, which inspired us to develop a multi-fidelity RL algorithm to train the FNN. To achieve this, We divide the DSE process into the low-fidelity (LF) phase and the high-fidelity (HF) phase.

\subsection{Low-fidelity Training with Model-based RL}
The LF phase is responsible for finding the promising area of the design space, which is supported by a large amount of data. To quickly obtain the CPI data, we adopt the analytical model proposed in \cite{jongerius2017analytic}. Given the design parameters and the profiling results of the target benchmarks, the model estimates the CPI based on the behavior abstraction of the processor, which takes about 0.1 ms per design. Interestingly, the analytical models are usually differentiable since they mainly consist of mathematical calculations. For non-differentiable operations like the lookup table, we can fit linear functions that strictly follow the trend of the table to acquire the gradients. Therefore, we can utilize the gradients of the analytical models to guide the DSE. 

Traditional model-based reinforcement learning (MBRL) \cite{moerland2023model} directly utilizes the gradients of the model to update neural network parameters, which requires the analytical model to accurately reflect the relationships between design parameters. 
However, the parameters with large gradients will always have higher priority to increase for conventional MBRL, requiring the analytical model to provide highly accurate gradients. Due to the non-linearity of the processor's analytical model and the fitting of non-differentiable components, the gradients of the model can only guarantee correct increasing or decreasing trends, but cannot reflect the importance of each design parameter. If traditional methods are used, parameters with larger gradients will have more opportunities to increase, but they may not bring satisfying benefits.
Therefore, we propose to only utilize the gradients to suggest the direction for updating. Specifically, we only allow the design parameters with negative gradients to be chosen for increasing at each step so that we can always take beneficial actions and increase the sampling efficiency. Further, to ensure the FNN finds the global optimum, we adopt an aggressive reward function design as shown in equation \ref{eq.1}.
\begin{equation}
    reward = IPC - IPC^* + \epsilon
\label{eq.1}
\end{equation}
where IPC is the reciprocal of CPI, $IPC^*$ is the observed highest IPC, and $\epsilon$ is a small value which ensures the optimal IPC can get a positive reward. In all our experiments, $\epsilon$ is 0.05.

\subsection{High-fidelity Training}
The adopted analytical model depends on bottleneck analysis to estimate IPC. However, its judgment of bottlenecks is not always accurate. As a result, when the analytical model determines that certain parameters cannot improve performance, we may discover in HF simulations that increasing these parameters still provides benefits. This allows us to make the most of the remaining area budget in the HF phase based on the LF results.

In the LF phase, the FNN collects the observed best designs and eventually converges to one of them. In order to transition from LF to HF, we first perform HF simulations on the converged design and a subset of the observed best designs. The obtained results are marked as $IPC_{h0}$ and $H$, respectively. The initial point of the HF phase is randomly sampled from $H$, then the FNN converged in the LF phase is used to decide the actions. Note that the actions in the HF phase are no longer restricted by the analytical model, therefore, the HF phase can explore the designs that are overlooked in the LF phase. 
Further, the reward function is modified to encourage the FNN to find better designs than in the LF phase and ensure a smooth transition between the two phases, which is shown in equation \ref{eq.2}.
\begin{equation}
    reward = IPC - IPC_{h0} + \epsilon
\label{eq.2}
\end{equation}

\begin{figure}[htbp]
\centerline{\includegraphics[width=0.8\columnwidth]{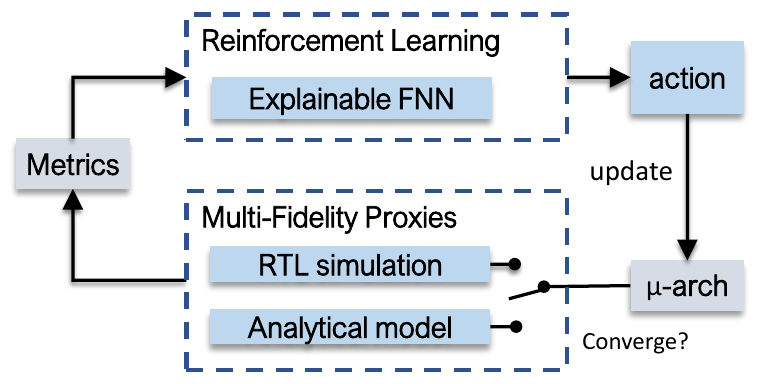}}
\caption{FNN with multi-fidelity RL.}
\label{fig.fuzzysys}
\end{figure}

Fig. \ref{fig.fuzzysys} shows the flow of the proposed multi-fidelity RL.


\section{Experiment}\label{sec.exp}
We conduct extensive experiments to validate the superiority of our method. In the HF phase, we use the Boom generator from Chipyard \cite{amid2020chipyard} to generate the RTL codes of the sampled designs, then we use VCS RTL simulator to obtain the CPI when running at 1GHz. We adopt McPAT \cite{li2009mcpat} to provide fast estimations for areas of the designs. 
    
\begin{table}[th]
    \centering
    {\fontsize{9pt}{10pt}\selectfont
    \caption{Design Space of $\mu$-arch.}
    \label{tab.designspace}
    \begin{tabular}{|c|c|}
      \hline 
      Parameters & Candidate values \\
      \hline
      \hline
      L1 Cache Set & 16, 32, 64 \\
      \hline
      L1 Cache Way & 2, 4, 8, 16 \\
      \hline
      L2 Cache Set & 128, 256, 512, 1024, 2048\\
      \hline
      L2 Cache Way & 2, 4, 8, 16 \\
      \hline
      nMSHR & 2, 4, 6, 8, 10 \\
      \hline
      Decode Width & 1, 2, 3, 4, 5 \\
      \hline
      ROB Entry & 32, 64, 96, 128, 160  \\
      \hline
      Mem FU & 1, 2 \\
      \hline
      Int FU & 1, 2, 3, 4, 5 \\
      \hline
      FP FU & 1, 2 \\
      \hline
      Issue Queue Entry & 2, 4, 8, 16, 24 \\
      \hline
    \end{tabular}\par}
\end{table}
To cover different types of applications, we select 6 benchmarks to evaluate CPI, i.e., dijkstra, matrix multiplication (mm), floating-point vector addition (fp-vvadd), quicksort, fast fourier transform (fft), string search (ss). Additionally, we increase the data sizes of these benchmarks to different extents to avoid the optimal results being concentrated on smaller $\mu$-arch designs.

The design space for the experiment is shown in Table \ref{tab.designspace}. We choose the design parameters that are jointly supported by the analytical model, Chipyard, and McPAT. 
The size of the whole design space is 3 million. Unlike previous works that build an offline dataset, we run all experiments online in the entire design space to simulate more realistic application scenarios.

\subsection{Evaluation of Application-Specific Usage}
To evaluate the effectiveness of our proposed method for application-specific design usage, we conduct DSE on each of the benchmarks. For each benchmark, we sample at least 500 points in the promising area, and the best one is considered the sampled optimal $\tilde{\text{{opt}}}$. 
Then, we can obtain the regrets, which is defined as the difference between the best result of DSE, denoted as $DSE_{\text{{best}}}$ and $\tilde{\text{{opt}}}$, i.e., 
\begin{equation}
 \text{{Regret}} = DSE_{\text{{best}}} - \tilde{\text{{opt}}}
\end{equation}

We compare the regret for the LF and HF results. The improvement of HF over LF is shown by the ratio of their regrets, i.e., 
\begin{equation}
\text{{Imp.}} = \frac{{\text{{Regret}}_{\text{{HF}}}}}{{\text{{Regret}}_{\text{{LF}}}}}
\end{equation}
As shown in Table \ref{tab.app-spec}, for all benchmarks, the HF significantly improves the results based on LF, and the results for mm, quicksort, and fft are almost $\tilde{\text{{opt}}}$, showing the effectiveness of our proposed multi-fidelity RL. 


\begin{table}[th]
    \centering
    \caption{Application-specific DSE results.}
    \begin{tabular}{ccccc}
      \toprule
       & area limit & LF regret & HF regret & Imp. \\
      \midrule
      dijkstra &10 $mm^2$  & 0.302 & 0.083  & 3.64 $\times$\\
      \hline 
      mm & 7.5 $mm^2$ & 0.020 & 0.007  & 2.86 $\times$\\
      \hline
      fp-vvadd & 6 $mm^2$ & 0.156 & 0.025  & 6.24 $\times$\\
      \hline
      quicksort & 7.5 $mm^2$ & 0.037 & 0.010  & 3.70 $\times$\\
      \hline
      fft & 8 $mm^2$ & 0.299 & 0.001  & 299.9 $\times$\\
      \hline
      ss & 6 $mm^2$ & 0.119 & 0.066  & 1.80 $\times$\\
      \bottomrule
    \end{tabular}
    \label{tab.app-spec}
\end{table}

\subsection{Evaluation of General-Purpose Usage}
\begin{figure}[htbp]
\centerline{\includegraphics[width=1.\columnwidth]{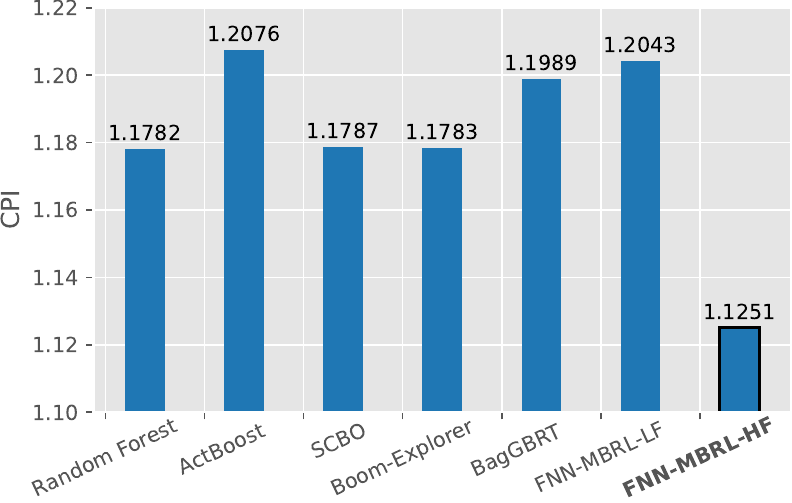}}
\caption{Comparison with baselines.}
\label{fig.baseline}
\end{figure}

To evaluate the effectiveness of our proposed method for general-purpose design usage, we further conduct DSE on the average of the results of all 6 benchmarks with an area constraint of 8 $mm^2$. We compare our method with the current state-of-the-art methods, e.g. Boom-Explorer\cite{bai2021boom}, BagGBRT\cite{wang2023high}, ActBoost\cite{li2016efficient}. We also include Scalable Constrained BO\cite{eriksson2021scalable}, a recent advance in BO which is competitive for high-dimensional constrained DSE problems. Further, a classic baseline Random Forest \cite{breiman2001random} is also included.
For all the baselines, we allow a budget of 10 HF simulations. To ensure all computation budgets are used on valid samples, the samples that violate the constraint are directly assigned a low reward and do not go through simulation, except for SCBO, which requires the invalid HF results to make inferences.
HF simulation takes around 2 hours to finish, which is the same as the LF training. For fair comparisons, we allow only 9 HF simulations for our method so that the running time is equal to the baselines. We run all methods with 5 different seeds and report the mean of the best CPI.
As shown in Fig. \ref{fig.baseline}, our proposed method significantly outperforms all baselines.

\subsection{Interpretability}
We also conduct several experiments to demonstrate the interpretability of our method.
The interpretability of the FNN is most directly reflected in the rule-based expression of the learning results.
To obtain the rules, we design a script that automatically translates the calculations of FNN into rules. We first map the matrix entries to the fuzzy values of the rules, then we prune the redundant parts of the rules to make it more clear for designers. To be detailed, a column of the matrix whose 1-norm is nearly 0 is considered redundant. Also, an antecedent item 'X' is redundant if 'X is high' and 'X is low' both claim a parameter can increase. We present some examples of the rules and briefly explain them.
\begin{itemize} 
    \item IF L1 is enough and FU is enough and decode is low, THEN decode can increase
    \item IF L1 is enough and FU is low THEN int can increase
    \item IF L1 is enough and ROB is enough and decode is enough and FU is low and IQ is low THEN IQ can increase
    \item IF L2 is low THEN ROB can increase
    \end{itemize}

These rules are decisions that provide high rewards, as recorded by the FNN during the training process. Since we used an analytical model to train the FNN, these rules are also a summary of the information provided by the analytical model. The first rule means that a relatively low decode width could be the bottleneck when L1 cache size is large and there are enough function units (FUs). As a larger L1 cache leads to less L1 miss and enough FUs allow higher throughput, the decoder should be able to handle more instructions, which is in line with common knowledge. 

The second rule claims that when L1 cache is sufficiently large and FUs are not enough, we can increase the number of integer units since less L1 miss causes a need for the FUs to process more instructions. The antecedents here do not include the decode width, as when the decode width is 1, the analytical model also identifies the integer unit as the bottleneck.

The third rule suggests that the issue queue (IQ) needs to increase when the L1 cache, reorder buffer (ROB) and decode width are large, but there are insufficient FUs, so the issue queue (IQ) needs to increase. To explain, if the first three components are large, it will lead to more instructions in flight. If there are not enough FUs, it is necessary to increase IQ entries to avoid stalls.

The last rule seems counter-intuitive due to the bias of the analytical model, which assumes that ROB stalls only occur due to L3 and DRAM access. Hence, when the L2 cache is large enough to hold all required data (ignoring the warmup phase), making the miss rate near 1, then ROB stalls are overlooked. Consequently, increasing the number of ROB entries is estimated to be unbeneficial.

\begin{figure}[t]
\centerline{\includegraphics[width=0.9\columnwidth]{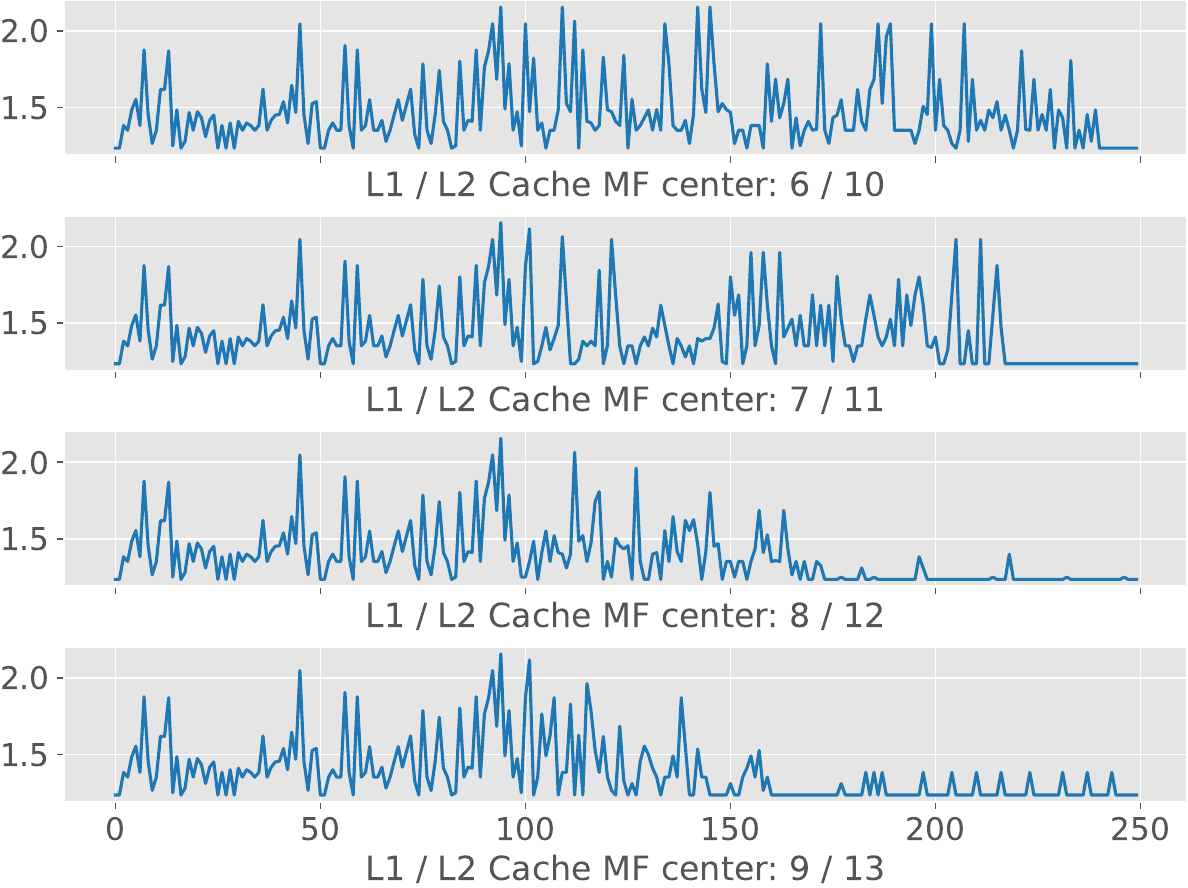}}
\caption{Comparison of different initialization.}
\label{fig.wise-init}
\end{figure}

FNN can provide such rules for all parameters, which makes it easy for designers to inspect the training results.
The last rule shows a limitation of the FNN that the quality of the summarized rules relies on the source of the information. If we want to form the perfect rule base, we need precise data and a comprehensive exploration of the data, which will result in a very slow convergence. Hence, the trade-off between interpretability and efficiency represents a principal challenge for researchers in the realm of explainable DSE. Despite this, our method has proven highly successful in making learning results interpretable.

Secondly, as we mentioned in Sec. \ref{sec.FNN}, designers can wisely decide the initial parameters to facilitate convergence. We largely increase the data size of dijkstra and run our method when L1 and L2 cache are differently initialized. As demonstrated in Fig. \ref{fig.wise-init}, higher MF centers achieve faster convergence. Importantly, all settings eventually converge, exhibiting the robustness of our method.

Last but not least, we show that designers can easily insert preferences into the FNN. We embed our preference for decode width 4 into the FNN rule base as described in Sec. 2.3 and conduct experiments on fp-vvadd, which originally converges to decode width 3. Changes of all $\mu$-arch parameters during training are shown in Fig. 7, where the blue line is the decode width, and grey lines are other parameters. Experiment results show that we successfully teach the decode width to reach 4. Unlike directly modifying parameters after sampling, we modify the knowledge of FNN, allowing FNN to generate the desired decision itself, which maintains the consistency of the model's learning process. 

\begin{figure}[tbp]
\centerline{\includegraphics[width=1.\columnwidth]{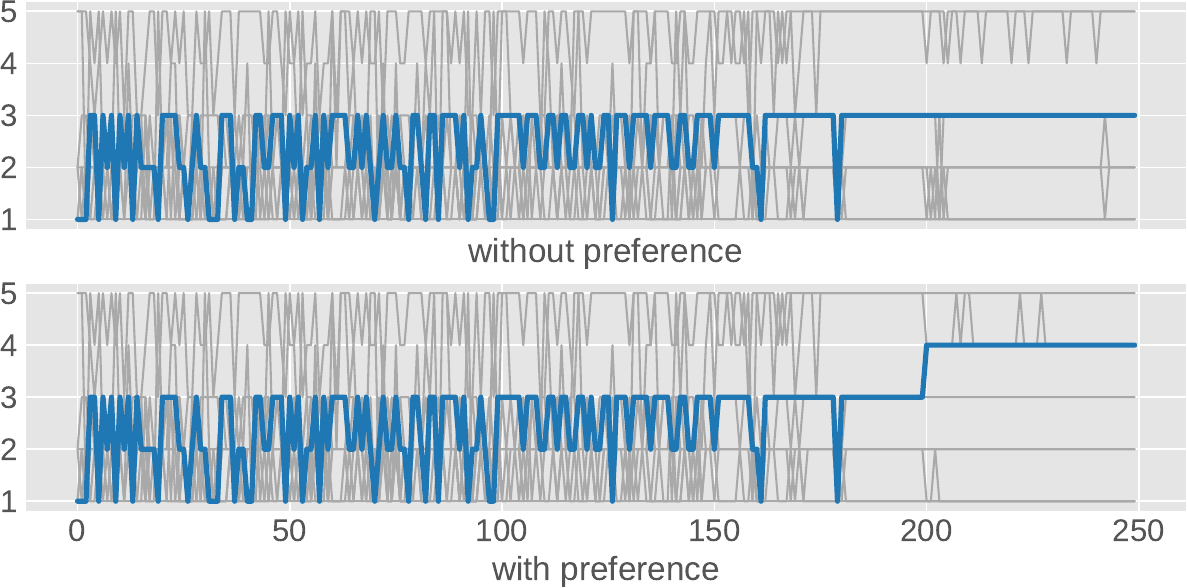}}
\caption{Embedding preference into FNN.}
\label{fig.prefer}
\end{figure}

\section{Conclusion}
In this work, we proposed to use FNN as the search engine for $\mu$-arch DSE, which makes the results explainable to human designers. The FNN is trained by our proposed multi-fidelity MBRL algorithm, which utilizes both the analytical model and the RTL simulator to ensure the accuracy of the results and reduce time consumption. The experiments show that our DSE framework achieves state-of-the-art results and provides good interpretability. 


\bibliographystyle{ACM-Reference-Format}
\bibliography{ref}


\begin{thebibliography}{20}


\ifx \showCODEN    \undefined \def \showCODEN     #1{\unskip}     \fi
\ifx \showDOI      \undefined \def \showDOI       #1{#1}\fi
\ifx \showISBNx    \undefined \def \showISBNx     #1{\unskip}     \fi
\ifx \showISBNxiii \undefined \def \showISBNxiii  #1{\unskip}     \fi
\ifx \showISSN     \undefined \def \showISSN      #1{\unskip}     \fi
\ifx \showLCCN     \undefined \def \showLCCN      #1{\unskip}     \fi
\ifx \shownote     \undefined \def \shownote      #1{#1}          \fi
\ifx \showarticletitle \undefined \def \showarticletitle #1{#1}   \fi
\ifx \showURL      \undefined \def \showURL       {\relax}        \fi
\providecommand\bibfield[2]{#2}
\providecommand\bibinfo[2]{#2}
\providecommand\natexlab[1]{#1}
\providecommand\showeprint[2][]{arXiv:#2}

\bibitem[Bai et~al\mbox{.}(2021)]%
        {bai2021boom}
\bibfield{author}{\bibinfo{person}{Chen Bai}, \bibinfo{person}{Qi Sun}, \bibinfo{person}{Jianwang Zhai}, \bibinfo{person}{Yuzhe Ma}, \bibinfo{person}{Bei Yu}, {and} \bibinfo{person}{Martin~DF Wong}.} \bibinfo{year}{2021}\natexlab{}.
\newblock \showarticletitle{BOOM-Explorer: RISC-V BOOM microarchitecture design space exploration framework}. In \bibinfo{booktitle}{\emph{2021 IEEE/ACM ICCAD}}. IEEE.
\newblock


\bibitem[Breiman(2001)]%
        {breiman2001random}
\bibfield{author}{\bibinfo{person}{Leo Breiman}.} \bibinfo{year}{2001}\natexlab{}.
\newblock \showarticletitle{Random forests}.
\newblock \bibinfo{journal}{\emph{Machine learning}}  \bibinfo{volume}{45} (\bibinfo{year}{2001}), \bibinfo{pages}{5--32}.
\newblock


\bibitem[Eriksson and Poloczek(2021)]%
        {eriksson2021scalable}
\bibfield{author}{\bibinfo{person}{David Eriksson} {and} \bibinfo{person}{Matthias Poloczek}.} \bibinfo{year}{2021}\natexlab{}.
\newblock \showarticletitle{Scalable constrained Bayesian optimization}. In \bibinfo{booktitle}{\emph{AISTATS}}. PMLR.
\newblock


\bibitem[et~al(2020)]%
        {amid2020chipyard}
\bibfield{author}{\bibinfo{person}{Alon~Amid et al}.} \bibinfo{year}{2020}\natexlab{}.
\newblock \showarticletitle{Chipyard: Integrated design, simulation, and implementation framework for custom socs}.
\newblock \bibinfo{journal}{\emph{IEEE Micro}} \bibinfo{volume}{40}, \bibinfo{number}{4} (\bibinfo{year}{2020}), \bibinfo{pages}{10--21}.
\newblock


\bibitem[Farhat et~al\mbox{.}(2016)]%
        {farhat2016fuzzy}
\bibfield{author}{\bibinfo{person}{Iqra Farhat}, \bibinfo{person}{Muhammad~Yasir Qadri}, \bibinfo{person}{Nadia~N Qadri}, {and} \bibinfo{person}{Jameel Ahmed}.} \bibinfo{year}{2016}\natexlab{}.
\newblock \showarticletitle{Fuzzy Logic-Based DSE Engine: Reconfiguration for Optimization of Multicore Architectures}.
\newblock \bibinfo{journal}{\emph{Journal of Circuits, Systems and Computers}} \bibinfo{volume}{25}, \bibinfo{number}{12} (\bibinfo{year}{2016}).
\newblock


\bibitem[{\"I}pek et~al\mbox{.}(2006)]%
        {ipek2006efficiently}
\bibfield{author}{\bibinfo{person}{Engin {\"I}pek}, \bibinfo{person}{Sally~A McKee}, \bibinfo{person}{Rich Caruana}, \bibinfo{person}{Bronis~R de Supinski}, {and} \bibinfo{person}{Martin Schulz}.} \bibinfo{year}{2006}\natexlab{}.
\newblock \showarticletitle{Efficiently exploring architectural design spaces via predictive modeling}.
\newblock \bibinfo{journal}{\emph{ACM SIGOPS Operating Systems Review}} \bibinfo{volume}{40}, \bibinfo{number}{5} (\bibinfo{year}{2006}).
\newblock


\bibitem[Jang(1993)]%
        {jang1993anfis}
\bibfield{author}{\bibinfo{person}{J-SR Jang}.} \bibinfo{year}{1993}\natexlab{}.
\newblock \showarticletitle{ANFIS: adaptive-network-based fuzzy inference system}.
\newblock \bibinfo{journal}{\emph{IEEE transactions on systems, man, and cybernetics}} \bibinfo{volume}{23}, \bibinfo{number}{3} (\bibinfo{year}{1993}).
\newblock


\bibitem[Jongerius et~al\mbox{.}(2017)]%
        {jongerius2017analytic}
\bibfield{author}{\bibinfo{person}{Rik Jongerius}, \bibinfo{person}{Andreea Anghel}, \bibinfo{person}{Gero Dittmann}, \bibinfo{person}{Giovanni Mariani}, \bibinfo{person}{Erik Vermij}, {and} \bibinfo{person}{Henk Corporaal}.} \bibinfo{year}{2017}\natexlab{}.
\newblock \showarticletitle{Analytic multi-core processor model for fast design-space exploration}.
\newblock \bibinfo{journal}{\emph{IEEE Trans. Comput.}} \bibinfo{volume}{67}, \bibinfo{number}{6} (\bibinfo{year}{2017}).
\newblock


\bibitem[Lee and Brooks(2007)]%
        {lee2007illustrative}
\bibfield{author}{\bibinfo{person}{Benjamin~C Lee} {and} \bibinfo{person}{David~M Brooks}.} \bibinfo{year}{2007}\natexlab{}.
\newblock \showarticletitle{Illustrative design space studies with microarchitectural regression models}. In \bibinfo{booktitle}{\emph{2007 IEEE 13th International Symposium on High Performance Computer Architecture}}. IEEE.
\newblock


\bibitem[Li et~al\mbox{.}(2016)]%
        {li2016efficient}
\bibfield{author}{\bibinfo{person}{Dandan Li}, \bibinfo{person}{Shuzhen Yao}, \bibinfo{person}{Yu-Hang Liu}, \bibinfo{person}{Senzhang Wang}, {and} \bibinfo{person}{Xian-He Sun}.} \bibinfo{year}{2016}\natexlab{}.
\newblock \showarticletitle{Efficient design space exploration via statistical sampling and AdaBoost learning}. In \bibinfo{booktitle}{\emph{Proceedings of the 53rd Annual Design Automation Conference}}.
\newblock


\bibitem[Li et~al\mbox{.}(2009)]%
        {li2009mcpat}
\bibfield{author}{\bibinfo{person}{Sheng Li}, \bibinfo{person}{Jung~Ho Ahn}, \bibinfo{person}{Richard~D Strong}, \bibinfo{person}{Jay~B Brockman}, \bibinfo{person}{Dean~M Tullsen}, {and} \bibinfo{person}{Norman~P Jouppi}.} \bibinfo{year}{2009}\natexlab{}.
\newblock \showarticletitle{McPAT: An integrated power, area, and timing modeling framework for multicore and manycore architectures}. In \bibinfo{booktitle}{\emph{MICRO 42}}.
\newblock


\bibitem[Lin and Lee(1994)]%
        {lin1994reinforcement}
\bibfield{author}{\bibinfo{person}{Chin-Teng Lin} {and} \bibinfo{person}{CS~George Lee}.} \bibinfo{year}{1994}\natexlab{}.
\newblock \showarticletitle{Reinforcement structure/parameter learning for neural-network-based fuzzy logic control systems}.
\newblock \bibinfo{journal}{\emph{IEEE Transactions on Fuzzy Systems}} \bibinfo{volume}{2}, \bibinfo{number}{1} (\bibinfo{year}{1994}).
\newblock


\bibitem[Moerland et~al\mbox{.}(2023)]%
        {moerland2023model}
\bibfield{author}{\bibinfo{person}{Thomas~M Moerland}, \bibinfo{person}{Joost Broekens}, \bibinfo{person}{Aske Plaat}, \bibinfo{person}{Catholijn~M Jonker}, {et~al\mbox{.}}} \bibinfo{year}{2023}\natexlab{}.
\newblock \showarticletitle{Model-based reinforcement learning: A survey}.
\newblock \bibinfo{journal}{\emph{Foundations and Trends{\textregistered} in Machine Learning}} \bibinfo{volume}{16}, \bibinfo{number}{1} (\bibinfo{year}{2023}), \bibinfo{pages}{1--118}.
\newblock


\bibitem[Sutton et~al\mbox{.}(1999)]%
        {sutton1999policy}
\bibfield{author}{\bibinfo{person}{Richard~S Sutton}, \bibinfo{person}{David McAllester}, \bibinfo{person}{Satinder Singh}, {and} \bibinfo{person}{Yishay Mansour}.} \bibinfo{year}{1999}\natexlab{}.
\newblock \showarticletitle{Policy gradient methods for reinforcement learning with function approximation}.
\newblock \bibinfo{journal}{\emph{Advances in neural information processing systems}}  \bibinfo{volume}{12} (\bibinfo{year}{1999}).
\newblock


\bibitem[Synopsys(2004)]%
        {synopsys2004verilog}
\bibfield{author}{\bibinfo{person}{VCS Synopsys}.} \bibinfo{year}{2004}\natexlab{}.
\newblock \showarticletitle{Verilog simulator}.
\newblock \bibinfo{journal}{\emph{Avaliable HTTP: http://www. synopsys. com/products/simulation/simulation. html}} (\bibinfo{year}{2004}).
\newblock


\bibitem[Takagi and Sugeno(1983)]%
        {takagi1983derivation}
\bibfield{author}{\bibinfo{person}{Tomohiro Takagi} {and} \bibinfo{person}{Michio Sugeno}.} \bibinfo{year}{1983}\natexlab{}.
\newblock \showarticletitle{Derivation of fuzzy control rules from human operator's control actions}.
\newblock \bibinfo{journal}{\emph{IFAC proceedings volumes}} \bibinfo{volume}{16}, \bibinfo{number}{13} (\bibinfo{year}{1983}).
\newblock


\bibitem[Wang et~al\mbox{.}(2023)]%
        {wang2023high}
\bibfield{author}{\bibinfo{person}{Duo Wang}, \bibinfo{person}{Mingyu Yan}, \bibinfo{person}{Yihan Teng}, \bibinfo{person}{Dengke Han}, \bibinfo{person}{Xiaochun Ye}, {and} \bibinfo{person}{Dongrui Fan}.} \bibinfo{year}{2023}\natexlab{}.
\newblock \showarticletitle{A High-accurate Multi-objective Ensemble Exploration Framework for Design Space of CPU Microarchitecture}. In \bibinfo{booktitle}{\emph{GLSVLSI 2023}}.
\newblock


\bibitem[Wilson et~al\mbox{.}(2016)]%
        {wilson2016deep}
\bibfield{author}{\bibinfo{person}{Andrew~Gordon Wilson}, \bibinfo{person}{Zhiting Hu}, \bibinfo{person}{Ruslan Salakhutdinov}, {and} \bibinfo{person}{Eric~P Xing}.} \bibinfo{year}{2016}\natexlab{}.
\newblock \showarticletitle{Deep kernel learning}. In \bibinfo{booktitle}{\emph{Artificial intelligence and statistics}}. PMLR, \bibinfo{pages}{370--378}.
\newblock


\bibitem[Zadeh(1988)]%
        {zadeh1988fuzzy}
\bibfield{author}{\bibinfo{person}{Lotfi~A Zadeh}.} \bibinfo{year}{1988}\natexlab{}.
\newblock \showarticletitle{Fuzzy logic}.
\newblock \bibinfo{journal}{\emph{Computer}} \bibinfo{volume}{21}, \bibinfo{number}{4} (\bibinfo{year}{1988}), \bibinfo{pages}{83--93}.
\newblock


\bibitem[Zeng et~al\mbox{.}(2010)]%
        {zeng2010framework}
\bibfield{author}{\bibinfo{person}{Zhipeng Zeng}, \bibinfo{person}{Reza Sedaghat}, {and} \bibinfo{person}{Anirban Sengupta}.} \bibinfo{year}{2010}\natexlab{}.
\newblock \showarticletitle{A framework for fast design space exploration using fuzzy search for VLSI computing architectures}. In \bibinfo{booktitle}{\emph{Proceedings of 2010 IEEE International Symposium on Circuits and Systems}}. IEEE.
\newblock


\end{thebibliography}


\end{document}